\documentclass[10pt,twocolumn,letterpaper]{article}

\usepackage{iccv}
\usepackage{times}
\usepackage{epsfig}
\usepackage{graphicx}
\usepackage{gensymb}
\usepackage{amsmath}
\usepackage{amssymb}
\usepackage{caption}
\usepackage{wrapfig}
\usepackage{paralist}
\usepackage{caption}
\usepackage{subcaption}

\usepackage[pagebackref=true,breaklinks=true,letterpaper=true,colorlinks,bookmarks=false]{hyperref}

\iccvfinalcopy 

\def\iccvPaperID{1327} 

\newcommand{\affmark}[1][*]{\textsuperscript{#1}}

\ificcvfinal\fi

\begin{document}

\title{\vspace{-6mm}Beyond Visual Attractiveness: Physically Plausible Single Image HDR Reconstruction for Spherical Panoramas}

\author{Wei Wei\affmark[1], Li Guan\affmark[2], Yue Liu\affmark[2], Hao Kang\affmark[2], Haoxiang Li\affmark[2], Ying Wu\affmark[1] and Gang Hua\affmark[2] \\
\affmark[1]Northwestern University, Evanston, IL, USA\\
\affmark[2]Wormpex AI Research, Bellevue, WA, USA \\
}

\maketitle

\newcommand{\li}[1]{{\color{magenta}{[{\bf\sf li:} #1]}}}
\newcommand{\yue}[1]{{\color{yellow}{[{\bf\sf yue:} #1]}}}
\newcommand{\hx}[1]{{\color{blue}{[{\bf\sf hx:} #1]}}}
\newcommand{\hao}[1]{{\color{green}{[{\bf\sf hao:} #1]}}}
\newcommand{\gang}[1]{{\color{cyan}{[{\bf\sf gang:} #1]}}}

\ificcvfinal\thispagestyle{empty}\fi

\begin{abstract}
   HDR reconstruction is an important task in computer vision with many industrial needs. The traditional approaches merge multiple exposure shots to generate HDRs that correspond to the physical quantity of illuminance of the scene. However, the tedious capturing process makes such multi-shot approaches inconvenient in practice. In contrast, recent single-shot methods predict a visually appealing HDR from a single LDR image through deep learning. But it is not clear whether the previously mentioned physical properties would still hold, without training the network to explicitly model them. In this paper, we introduce the physical illuminance constraints to our single-shot HDR reconstruction framework, with a focus on spherical panoramas. By the proposed physical regularization, our method can generate HDRs which are not only visually appealing but also physically plausible. For evaluation, we collect a large dataset of LDR and HDR images with ground truth illuminance measures. Extensive experiments show that our HDR images not only maintain high visual quality but also top all baseline methods in illuminance prediction accuracy.
\end{abstract}

\section{Introduction} \label{sec:intro}
\begin{figure}
    \centering
    \includegraphics[width=1\linewidth]{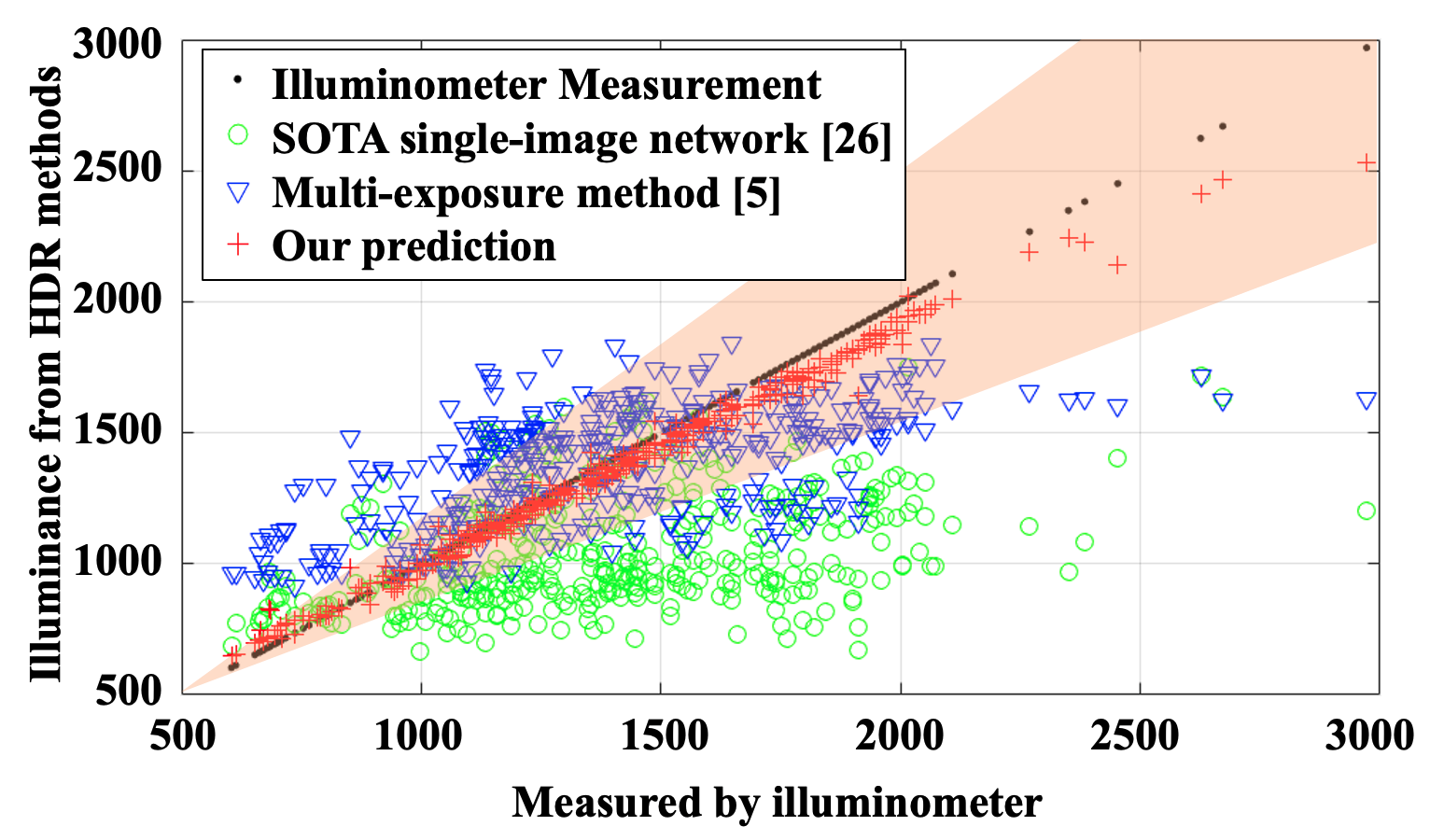}
    \caption{The physical illuminance metric performance in our practical dataset captured in multiple shopping stores. Every data point corresponds to a generated HDR image in the physical world. Black: illuminometer readout at the imaging location, as the ground truth. Orange shaded cone: $\pm$25\% absolute error confidence interval around the ground truth. Green: estimations from state-of-the-art single-shot HDR reconstruction method \cite{liu2020single}. Blue: estimations from classical multi-shot HDR reconstruction method \cite{debevec2008recovering} implemented with \textit{Photosphere} \cite{ward2016anyhere}. Red: estimations from our method. Our illuminance estimations mostly lie within the confidence interval.}
    \label{fig:fig1}
\end{figure}

When a High Dynamic Range (HDR) image correctly captures the wide luminance variation within a scene \cite{debevec2008recovering}, it could serve as a radiance map to support photometric applications in illuminating engineering \cite{moeck2006illuminance}, as well as computer vision applications like image rendering/compositing \cite{hold2017deep,legendre2019deeplight}, image relighting \cite{wen2003face}, \etc. We call such HDR \emph{physically plausible}, as it provides faithful illuminance\footnote{\emph{illuminance} refers to the amount of light falling onto a given surface area. It can be objectively measured by an illuminometer.} estimation that agrees with the readout from an illuminometer at the imaging location.

To reconstruct a physically plausible HDR, early methods \cite{debevec2008recovering} rely on merging multiple Low Dynamic Range (LDR) images that capture the same scene but with different exposure settings. By recovering the camera response function, they derive that the pixel values of the reconstructed HDR are proportional to the true physical quantity of luminance in the scene. Specifically, researchers in Optics \cite{inanici2006evaluation,pierson2019tutorial} have shown that HDRs reconstructed by such multi-shot method captured with certain camera equipment (180\degree fisheye lens or 360\degree omnidirectional cameras) can correspond to the real illuminance value with reasonable precision (within 10\% error margin). However, these results are built upon strict static-scene assumptions and expensive color and vignetting calibration procedures, which significantly limit the applicability of this technique in practice.

As a comparison, reconstructing HDR from only a single-shot can alleviate this issue. Recently, single-shot HDR reconstruction has achieved remarkable performance by learning a deep neural network mapping from an LDR image to its corresponding HDR ground truth \cite{endo2017deep,lee2018deeprecursive,lee2018deepchain,eilertsen2017hdr,zhang2017learning,marnerides2018expandnet,eilertsen2017hdr,Marcel:2020:LDRHDR,liu2020single}. They adopt various image-level priors with different networks to achieve visually appealing results. Despite their attractive HDR outputs, these methods are not trained to preserve scene physical illuminance. Thus, it is not clear whether their reconstructed HDRs are physically plausible and the useful physical property which is of interest for illuminating engineering is preserved.

This motivates us to study \textbf{reconstructing a physically plausible HDR from a single-shot}. In this work, we propose a novel idea that is to train the deep network with explicit physical prior modeling, which regularizes the highly-unconstrained reconstruction process toward a physically correct direction. This clearly distinguishes our method from the state-of-the-art learning-based single-shot HDR reconstructions methods which focus on image-level prior modeling. Specifically, we introduce an illuminance regularization term that forces the illuminance value derived by the HDR prediction to be close to the ground truth illuminance measured by an illuminometer. To derive the illuminance value from the HDR prediction, we need to perform a surface integration of all the pixel luminance values (computed from pixel intensities) over the hemispheric field-of-view (FOV) captured in the HDR image. Therefore, we focus on reconstructing physically plausible HDR from single-shot for spherical cameras (\eg, fisheye or 360\degree cameras), which are nowadays commonly available and frequently used in applications like autonomous driving and mixed reality. Note that although focused on the spherical images, our method can be further generalized to reconstructing physically plausible HDR for an arbitrary image with limited FOV, since if the physical plausibility of the spherical HDR generated by our method is verified, its cropped counterpart with limited FOV can be naturally utilized as the training ground truth for learning the HDR reconstruction for general images.

In addition, we discover that for many single-shot HDR reconstruction methods, the HDR outputs of the photographs capturing the same scene but with different exposure times may not be consistent. This also violates the common sense that the physically plausible HDRs should be identical if the scene remains the same. To solve this problem, our idea is to adopt the camera exposure parameter, which is usually available as part of the EXIF nowadays, as an additional network input. Note that it is not mandatory to provide such information in our framework. But when provided with the auxiliary exposure input, it would guide the learning process to be more robust.

Subjectively, our new framework has three advantages. 1) Plausibility: by explicitly modeling the physical prior into the learning process, we can generate physically plausible HDR outputs. As illustrated in Figure \ref{fig:fig1}, our HDR derives more accurate illuminance estimations compared to the state-of-the-art single-shot HDR reconstruction method \cite{liu2020single}. 2) Consistency: by encoding the additional exposure information, we can generate consistent HDR outputs for photographs capturing the same scene but with different exposure settings. 3) Convenience: our method, which only requires a single-shot, is more convenient than multi-shot methods. Multi-shot methods can achieve physically plausible HDRs but in the cost of strictly controlled lab environment with perfect static scenes and calibrated systems, limiting its applicability. On the other hand, our method bypasses these burdens and generates the most physically plausible HDRs compared to the multi-shot and single-shot baselines in our practical dataset, as shown in Figure \ref{fig:fig1}.

Overall, our contributions are three-fold:
\begin{compactenum}
    \item We propose a novel learning-based single-shot HDR reconstruction network that can generate HDR with the advantage of physical plausibility and consistency. 
    \item We introduce two new objective metrics to evaluate the performance of physical plausibility and consistency on the investigated task.
    \item We collect a new dataset which contains $\sim$8k HDR panoramas with measured ground truth illuminance, captured under various exposure times, lighting and scene conditions. We will release the dataset and the code for further research purposes in this community.
\end{compactenum}

\vspace{-1mm}

\section{Related Work} \label{sec:related}
\vspace{-1mm}
\subsection{HDR Reconstruction}

\noindent
\textbf{Single-shot HDR Reconstruction.} 
Reconstructing HDR from only a single-shot is very challenging, since the quantization and saturation of camera sensors cause irreversible information loss in underexposed and overexposed regions. Some methods are based on designing specific sophisticated camera systems or optical architectures, like modulo camera \cite{zhao2015unbounded} or beam-splitters \cite{tocci2011versatile}, and achieve pleasing results. Nevertheless, their equipment are custom-grade and expensive. In this work, we aim at solving this task for consumer-grade cameras and study it from an algorithm perspective.

To solve this ill-posed problem, how to utilize domain knowledge to compensate for the missed information caused by consumer-grade camera sensors is the key. The general insight is to learn from data \cite{endo2017deep,lee2018deeprecursive,lee2018deepchain,eilertsen2017hdr,zhang2017learning,marnerides2018expandnet,eilertsen2017hdr,Marcel:2020:LDRHDR,liu2020single}. By learning the domain knowledge from data, either through implicitly modeling the deep network structure or through explicitly modeling image priors, these methods can generate visually appealing HDRs. However, whether or not the results are physically plausible is not explored.

Among these methods, one way is to infer bracketed LDR images with different exposure times from a single LDR image input and then merge the inferred bracket into HDR by multi-shot methods. Specifically, \cite{endo2017deep} applies 3D CNNs, \cite{lee2018deeprecursive} utilizes GAN and \cite{lee2018deepchain} adopts deep chain network in the bracket inferring step. 

The other way is to directly learn an LDR-to-HDR mapping. \cite{eilertsen2017hdr} designs an autoencoder structure with a skip-connection scheme to recover saturated information in over-exposed regions. \cite{zhang2017learning} also proposes a deep autoencoder framework to regress HDR from LDR outdoor panoramas to predict outdoor lighting. \cite{marnerides2018expandnet} designs an end-to-end CNN architecture named ExpandNet with parallel branches and feature fusion. Two recent works make improvements over the method of \cite{eilertsen2017hdr}. \cite{Marcel:2020:LDRHDR} introduces a feature masking technique to alleviate the influence of the features from the saturated areas while still preserving unsaturated areas. \cite{liu2020single} models the HDR-to-LDR formation pipeline by three successive modules, \ie, dynamic range clipping, non-linear mapping and quantization, and decomposes the reversed pipeline into three subnetworks. Image-level regularization terms like Total-Variation loss and perceptual loss are adopted to generate smooth and vivid HDR outputs.

From above, the aforementioned single-shot HDR reconstruction methods make great efforts in reconstructing visually appealing HDRs, but lacking consideration on the physical property. Instead, we adopt the learning framework and further introduce the physical prior that makes the HDR generations more physically plausible, while not compromising its visual effects.

\noindent
\textbf{Multi-shot HDR Reconstruction.} Beyond reconstruct HDR from the single-shot which needs external training data, one common way to generate HDR is by merging multiple LDR images, captured using exposure bracketing, a technique of taking several shots of the same subject using different camera settings. As a representative work, \cite{debevec2008recovering} fuses exposure bracketing into a single HDR image whose pixel intensities are proportional to the real radiance values of the scene, with the recovered camera response function. 
Excellent results can be achieved for static scenes with this approach. However, when it comes to scenarios with dynamic motion (camera motion or object motion), the results are not satisfactory with ghosting artifacts. To eliminate the artifacts, multiple alignment methods are proposed, \eg, by HDR stitching \cite{kang2003high}, deghosting \cite{gallo2009artifact,hu2013hdr}, optical flow \cite{kalantari2017deep}, patch-based optimization \cite{sen2012robust} and deep networks \cite{wu2018deep,yan2019attention,yan2020deep}. 
Still, it is less convenient than directly generating HDR from a single-shot. In contrast, our method shares the physical plausibility of multi-shot methods, at the same time, bypasses the strict conditions and complex post-processing procedures.

\subsection{HDR and Illuminance Measurement}\label{HDR and Illuminance Measurement}

Capturing wide luminous variations with HDR photography is well studied in Photometry \cite{inanici2009introduction}. In \cite{inanici2010evalution}, it has been shown with laboratory experiments that the pixel intensities of HDR photographs generated by the multi-shot method \cite{debevec2008recovering} are aligned with physical luminance measurements up to a scale. The scaling factor depends on the configurations of the camera equipment and sometimes the specific scene, and needs to be calibrated through linear regression.


Theoretically, illuminance value (typically expressed in \textit{lux}) can be derived from the integration of luminance values in hemispherical fisheye images, where each differential area corresponds to the original area multiplied by the cosine of the polar angle \cite{ward1998rendering}. 

Deriving accurate illuminance values from HDR images demands strong assumptions (\ie, static scene) and rigorous calibration processes. For example, the vignetting effect (\ie, the brightness decrease from the center of a picture toward its periphery \cite{jacobs2007determining}) needs to be corrected to get accurate brightness distribution over the entire picture. This requires professional equipment (\eg, semi-circular platform \cite{cauwerts2012comparison}). Moreover, as mentioned earlier, a camera-depended scaling factor needs to be estimated such that the re-scaled HDR image can correspond to the luminance map correctly. 

By leveraging the deep neural networks, our method eases the burden of the complicated calibration processes of the multi-shot methods while maintaining the physically plausible property of the output HDR. 

\section{Model} \label{sec:model}
\begin{figure*}[!t]
    \centering
    \includegraphics[width=0.9\linewidth]{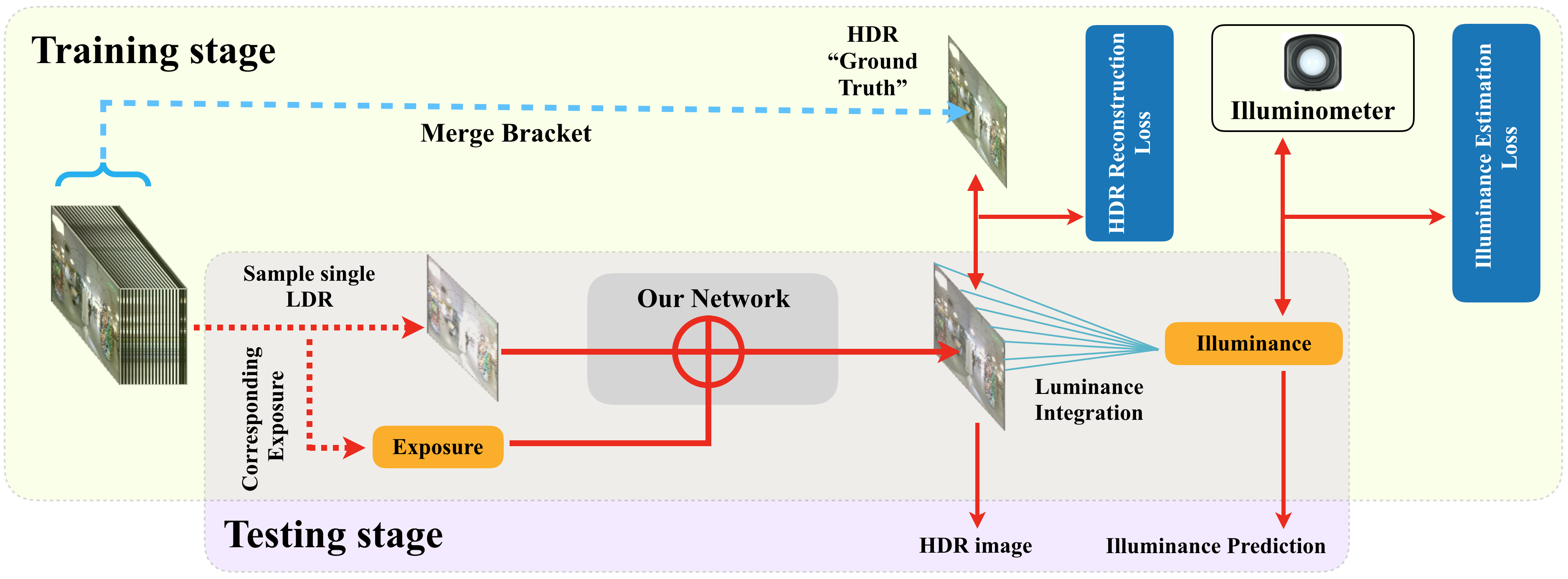}
    \caption{Overall framework of our network. In testing stage, our network takes as inputs the LDR image and (optional) exposure value to reconstruct an HDR image. The illuminance value of the scene can then be derived from the HDR image. In training, the exposure-conditioned HDR reconstruction process is supervised by ``ground-truth" HDR generated from the multi-shot captured at the same location, as well as the illuminometer reading at the time of capture.}
    \label{fig:network}
\end{figure*}

We first introduce how to derive the illuminance value from an HDR image in Section \ref{sec3.1}. Then we present our illuminance guided HDR reconstruction neural networks in Section \ref{sec3.2}. Last, we introduce the exposure-conditioned design for more robust and consistent HDR reconstruction in Section \ref{sec3.3}. The overall framework is shown in Figure \ref{fig:network}.

\subsection{Deriving Illuminance from HDR}\label{sec3.1}

Shown as the luminance integration step in Figure~\ref{fig:network}, the illuminance values can be deterministically derived from rigorous HDR images. First, a luminance map $L$ can be calculated (in $cd/m^2$) by the following linear transformation from R, G, and B channels of the HDR image \cite{pierson2019tutorial}:
\begin{equation}
    \vspace{-1mm}
    L = 179 \times( 0.2126\cdot R + 0.7152\cdot G + 0.0722\cdot B),
    \label{eq:rgb2luminance}
\end{equation}

The input R, G, B channels in Eq. \ref{eq:rgb2luminance} are in Radiance RGBE format, which can be produced with Photosphere \cite{ward2016anyhere}, a multi-shot based HDR generation software. This linear transformation has been validated by many Photometry papers \cite{inanici2006evaluation, inanici2009introduction, inanici2010evalution, pierson2019tutorial} comparing the pixel values in resulting HDR photographs to the physical quantity of illuminance (measured by illuminometers). Due to the page limit, we refer readers to \cite{pierson2019tutorial} for more details.

The illuminance of the imaging location, measuring the amount of light entering that location, can then be derived by integration of incident light energy of the luminance map multiplied by the cosine of solid angle. In our platform, an illuminometer normally only measures incident light from a hemisphere, we therefore only need to integrate the corresponding hemisphere as well:
\begin{equation}
    \hat{\mathcal{I}} = \int_{0}^{2\pi} \int_{0}^{\pi/2} L(\theta,\phi)\sin\theta\cos\theta d\theta d\phi,
    \label{eq:integration}
\end{equation}
where $\hat{\mathcal{I}}$ is the derived illuminance value estimated from HDR luminance map, $\theta,\phi$ are the zenith and azimuth angle of spherical coordinate system.

\subsection{Illuminance Guided HDR Reconstruction}\label{sec3.2}

We can derive illuminance value from an appropriate HDR image through a differentiable process. We leverage the real illuminance value measured by an illuminometer as additional supervision in HDR reconstruction, \ie,
\begin{equation}
    \mathcal{L}_{illuminance} = ||\hat{\mathcal{I}} - \mathcal{I}_{gt}||_2^2,
    \label{eq: loss_illuminance}
\end{equation}
where $\hat{\mathcal{I}}$ is the derived illuminance value in Equation \ref{eq:integration}, and $\mathcal{I}_{gt}$ is the ground truth illuminance value measured by the illuminometer.

\begin{figure}[!t]
    \centering
    \includegraphics[width=\linewidth]{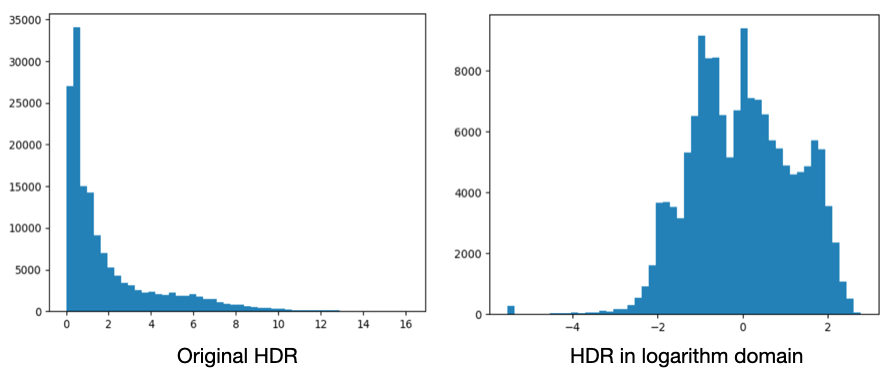}
    \vspace{-8mm}
    \caption{Comparison of the intensity histograms of the HDR and its logarithmic version.}
    \label{fig:log}
    \vspace{-4mm}
\end{figure}

We build our LDR-to-HDR neural networks upon a recent state-of-the-art single-shot HDR reconstruction pipeline~\cite{liu2020single}. The pipeline is implemented as a network composite of three sub-networks --- \textit{dequantization-net}, \textit{linearization-net}, and \textit{hallucination-net} to simulate the reversed camera imaging process. To train the network, several losses are combined:
\begin{equation}
    \mathcal{L}_{HDR} = ||\log(\hat{H})-\log(H_{gt})||_2^2 + \lambda_{TV}||\hat{H}||_{TV} + \lambda_p\mathcal{L}_p,
    \label{eq: loss_hdr}
\end{equation}
where $\hat{H}$ is the predicted HDR image by the network, and $H_{gt}$ is the ground truth HDR image. Since the HDR intensities follow a long-tailed distribution while its logarithm version obeys a Gaussian-like distribution, as shown in Figure \ref{fig:log}, we use L2 loss in the logarithm domain for more stable optimization results.

Other widely used terms are added into $\mathcal{L}_{HDR}$ as well, including a Total Variation \cite{rudin1992nonlinear} regularization term of the predicted HDR image to encourage spatial smoothness, and a perceptual loss term $\mathcal{L}_p$ ~\cite{johnson2016perceptual} with VGG-16 \cite{simonyan2014very} features to boost the vividness of the reconstructed HDR image.

To summarize, the final loss function to train the whole pipeline is as follows:
\begin{equation}
    \mathcal{L} = \mathcal{L}_{HDR} + \lambda \mathcal{L}_{illuminance}.
    \label{eq:loss_total}
\end{equation}
Despite its conciseness, the illuminance loss term could restrict the highly unconstrained solution space and regularize the HDR prediction to be physically plausible.

\subsection{Exposure-Conditioned HDR Reconstruction}\label{sec3.3}
On top of state-of-the-art HDR reconstruction neural networks, we design additional channels taking the exposure value of input LDR. The auxiliary input can potentially help the neural networks better handle the over-exposed and under-exposed areas to make more accurate HDR predictions. In practice, this information is usually available as part of the EXIF file for most consumer-grade cameras. 

More specifically, the exposure values are discretized into a 20-dimensional one-hot vector corresponding to a range from 5ms to 100ms with a step size of 5ms. As shown in Figure~\ref{fig:network}, the encoded exposure is then used as additional inputs to the HDR reconstruction neural networks. The specific way to incorporate this information varied by design. In our method, we concatenate the embedding \cite{mikolov2013distributed} of the one-hot representation of exposure value with the hidden feature layer with the smallest resolution in the U-Net structure. The HDR output is thus conditioned on the given exposure value with the input LDR image. The details of the network structure are explained in supplementary material.


\section{Experiments} \label{sec:exp}
\begin{table*}[!htbp]
\centering
\vspace{-4mm}
{\small
\begin{tabular}{|c|c|c|c|c|c|c|} 
 \hline
 Store ID & \scriptsize{\# of LDRs} & \scriptsize{\# of HDR and illuminance} & OLSE of scaling factor & Calibrated 25\% Acc & Calibrated 10\% Acc\\ 
 \hline\hline
 0001 &  52500 & 2625 & 1.4514 &70.1\% & 27.2\%\\
 0002 &  20640 & 1032 & 1.8207 &78.8\% & 32.9\%\\ 
 0003 &  24160 & 1208 & 1.4905 &62.5\% & 25.1\%\\ 
 0004 &  17860 & 893 & 1.4626 &68.3\% & 30.9\%\\ 
 0005 &  17280 & 864 & 1.5645 &59.3\% & 25.9\%\\
 0006 &  12920 & 646 & 1.7315 &57.6\% & 21.7\%\\ 
 0007 &  9480  & 474 & 1.6403 &82.1\% & 46.4\%\\ 
 0008 &  13860 & 693 & 1.4042 &54.5\% & 26.6\%\\
 \hline
\end{tabular}
}
\vspace{-2mm}
\caption{\textbf{Dataset HDRPano-I.} We collect a dataset of HDR panoramas, and measure their corresponding illuminance. The dataset is captured by 8 different imaging robots from 8 stores. We also estimate a per-robot HDR-to-illumiance scaling factor. The variation indicates it is a necessary step to evaluate physical illuminance property across multiple imaging devices.}
\label{table:summary}
\end{table*}

In this section, we first introduce the new dataset we collected and the evaluation protocol; then we compare our results against state-of-the-art methods under both same-domain and cross-domain settings; finally, we report results of ablation studies to provide more insights on different components of the proposed method.



\noindent
\subsection{HDRPano-I Dataset}

As far as we know, there is no publicly available large-scale HDR dataset with illuminance measures, thus we do not report results on these benchmark datasets \cite{gardner2017learning,hold2019deep}. Instead, we use a robot platform to collect a new dataset across 8 retail stores. The robot is programmed with varying camera shutter speeds (at 20 stops, ranging from 5\textit{ms} to 100\textit{ms}), capturing locations (8435 locations in total) at different time-of-day to collect images and measure illuminance. Both the 360-degree panorama cameras (dual fisheye-lens camera~\cite{sony}) and the illuminometer~\cite{maxim} (looking upward) are mounted at human eye-level height. The camera white balance and contrast setting are fixed to minimize the color shift problem. We summarize the details of the collected data in Table~\ref{table:summary}.


We generate the HDR from 20 captured LDR panoramas with \emph{Photosphere} ~\cite{ward2016anyhere} software. Note that due to the real-world manufacturing variations of both the robot and the cameras, the camera response curves across cameras are not the same. Also, HDR images generated by this multi-shot method~\cite{debevec2008recovering} need to be calibrated with a scaling factor (see Section \ref{HDR and Illuminance Measurement}). For each device, the scaling factor is obtained from an Optimal Least Square Estimation (OLSE)~\cite{debevec2008recovering}, as shown in Table~\ref{table:summary}. All our training and testing are conducted on the calibrated HDR images. 

We integrate the upper hemisphere of the HDR panoramas to estimate the illuminance value~\cite{inanici2010evalution}. As shown in Table~\ref{table:summary}, compared with the illuminometer readings, the illuminance estimated from calibrated HDR images still has a gap. Here $25\%$ accuracy represents the percentage of the predictions that are within $25\%$ error margin. We argue that it is difficult to obtain physically plausible HDR images in practice that even after carefully following the process 
~\cite{pierson2019tutorial}, the luminance of obtained HDR images is still far from perfect. Limitations in practice leading to the gap include moving objectives during the capturing session and missed camera lens vignetting calibration. The latter is a labor-intensive procedure impractical for large-scale deployment. 

Hence although the HDR images provided in our dataset are visually appealing, they are still not real ground-truth HDR images, but are representations of considerable efforts in real-world applications. Our method by design takes the illuminance readings as supervision and alleviates the negative impacts from dynamic objects in scenes and being lack of photometric calibrations.

\vspace{-1mm}
\subsection{Evaluation Metrics and Protocol}
\vspace{-1mm}

We adopt three metrics for evaluating the performance of single-shot HDR reconstruction. The first one is the commonly used HDR-VDP-2 (Visual Difference Predictor for HDR images) \cite{mantiuk2011hdr} for this task, which compares a pair of images (reference and reconstructed) and predicts whether differences between two images are visible to the human observer. The HDR-VDP-2 works within the complete range of luminance the human eye can see. The range is from 0 to 100, with a higher metric representing less visible difference.

The second metric is calculated as the mean value of the standard deviation map of HDRs reconstructed from each LDR image in the exposure bracket. By physical correctness, LDR images captured in the same session should construct the same HDR. The smaller the \emph{Mean Std} metric is, the better consistency of HDR images is.

More importantly, our third metric is the illuminance estimation accuracy from the reconstructed HDR image. Two values are reported using this metric: 25\% accuracy and 10\% accuracy, which denote the accuracy of derived illuminance value from reconstructed HDR within the absolute error of 25\% and 10\% of the true illuminance measured by illuminometer. No prior work has introduced this metric as well as the \emph{Mean Std} metric to the investigated task.

We conduct the experiments in two settings: same-domain and cross-domain. In the same-domain study, the training, validation and testing data share the data captured from the same stores, split by capture locations. While for cross-domain study, we evaluate the performance of training from the data in several stores and testing in the rest without fine-tuning, to verify whether our trained model can generalize to the new domain.

\subsection{Implementation Details}
We implement our method with PyTorch \cite{paszke2017automatic}, with two GPUs of RTX 2080. We use the gradient method of Adam \cite{kingma2014adam} to train the network. The initial learning rate is 1e-4. We train 10 epochs in total with the learning rate decayed to 1/10 at the 5-th and 8-th epoch. $\lambda_{TV}$ and $\lambda_p$ in Equation \ref{eq: loss_hdr} are set as 0.1 and 0.001. $\lambda$ in Equation \ref{eq:loss_total} is tuned as 1.

\subsection{Results of Same-domain Study}

\begin{table}[!t]
\centering
\vspace{-4mm}
{\small
\begin{tabular}{|c | c| c| c| c|} 
 \hline
  Methods & \scriptsize{HDR-VDP}$\uparrow$ & \scriptsize{Mean Std}$\downarrow$ & \scriptsize{25\% Acc}$\uparrow$ & \scriptsize{10\% Acc}$\uparrow$ \\ 
 \hline\hline
 \multicolumn{5}{|c|}{\textbf{Multi-shot method} \cite{ward2016anyhere}}\\
 \hline
 \scriptsize{Before Calibration} & --- & --- & 30.75\% & 10.00\% \\
 \scriptsize{After Calibration} & --- & --- & 74.75\% & 31.25\% \\
 \hline
 \multicolumn{5}{|c|}{\textbf{Single-shot methods}}\\
 \hline
 \small{HDRCNN}\cite{eilertsen2017hdr} & 64.4616 & 0.2599 & 7.26\% & 2.18\% \\
 \small{DrTMO}\cite{endo2017deep} & 66.3621 & 0.2162 & 8.13\% & 2.47\%\\ 
 \small{RevPipeline} \cite{liu2020single} & 70.7749 & 0.9499 & 30.66\% & 12.24\% \\ 
 Ours & 72.1717 & \textbf{0.0959} & \textbf{99.98\%} & \textbf{97.95\%} \\ 
 \hline
\end{tabular}
}
\vspace{-2mm}
\caption{Same-domain quantitative comparison on HDR images with existing methods.}
\label{table:same}
\end{table}

For this experiment, we use the data of StoreID `0001' and `0002' with 3657 HDRs in total. We randomly split the data into 3000 captured locations for training, 257 for validation and 400 for testing. The same procedure is repeated five times to erase the influence of splitting randomness. 

We compare with a series of state-of-the-art single-shot HDR reconstruction methods: HDRCNN \cite{eilertsen2017hdr}, DrTMO \cite{endo2017deep}, and RevPipeline \cite{liu2020single}, as well as multi-shot baseline \cite{ward2016anyhere}. We train HDRCNN and DrTMO with our dataset from scratch. For RevPipeline, they train three subnetworks and finally finetune in an end-to-end manner. Since we do not explicitly define the ground truth of intermediate output, we finetune their pretrained model with our data.

As shown in Table~\ref{table:same}, our method significantly outperformed others in the illuminance estimation due to the supervision from illuminance readings. By leveraging the deep networks, our prediction is robust to factors such as moving objects and in-accurate photometric calibration and performed even better than the ``ground-truth'' HDR images (``After Calibration'' in Table \ref{table:same}) obtained through the multi-shot method. With the exposure-conditioned design in our deep neural networks, our method also outputs more consistent HDR images, evidenced by the \emph{Mean Std} metric. The HDR-VDP metric indicates we can achieve comparable results to the state-of-the-art methods in terms of HDR visual quality. We adopt this metric only to verify the reconstructed HDR results of our method are realistic. 

\begin{figure*}
    \vspace{-4mm}
    \centering
    \begin{subfigure}{\textwidth}
    \centering
    \includegraphics[width=0.9\linewidth]{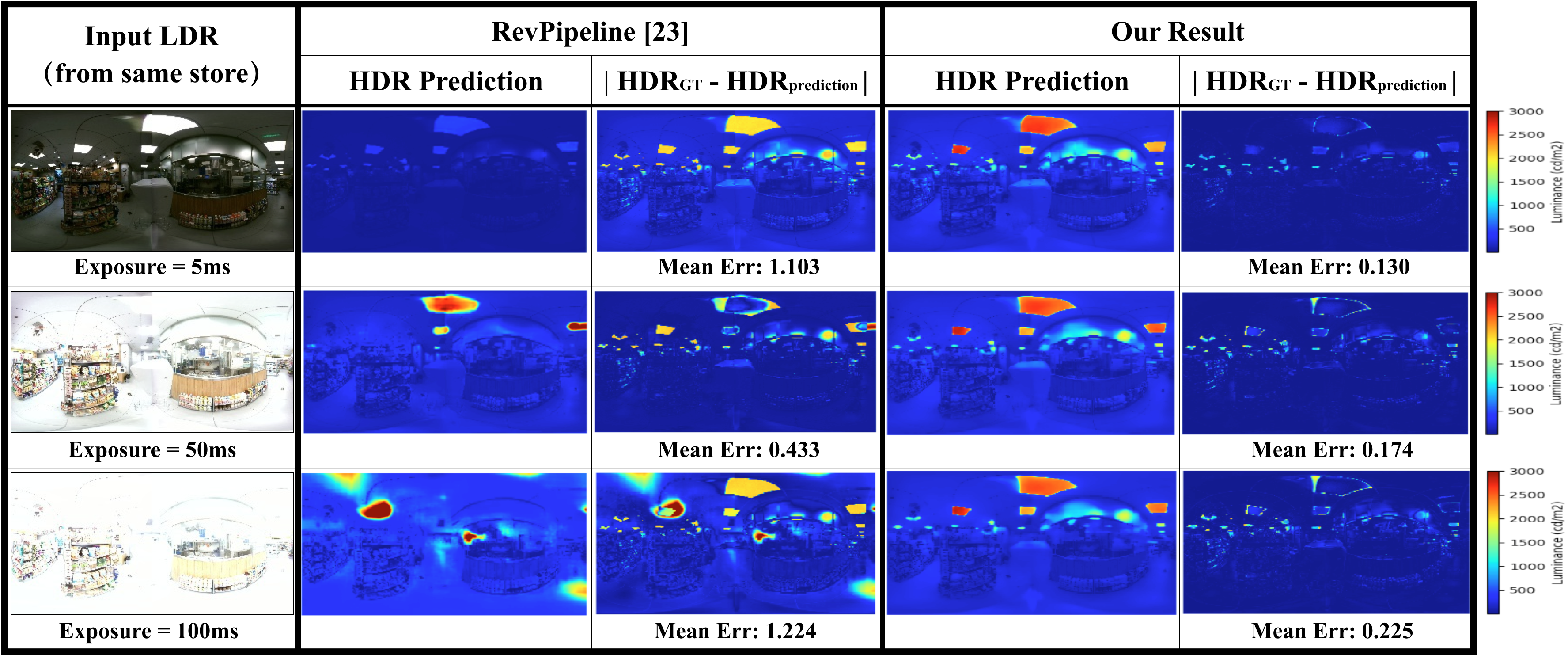}
    \caption{Visualization of same-domain study.}
    \end{subfigure}
    \hfill
    \begin{subfigure}{\textwidth}
    \centering
    \includegraphics[width=0.9\linewidth]{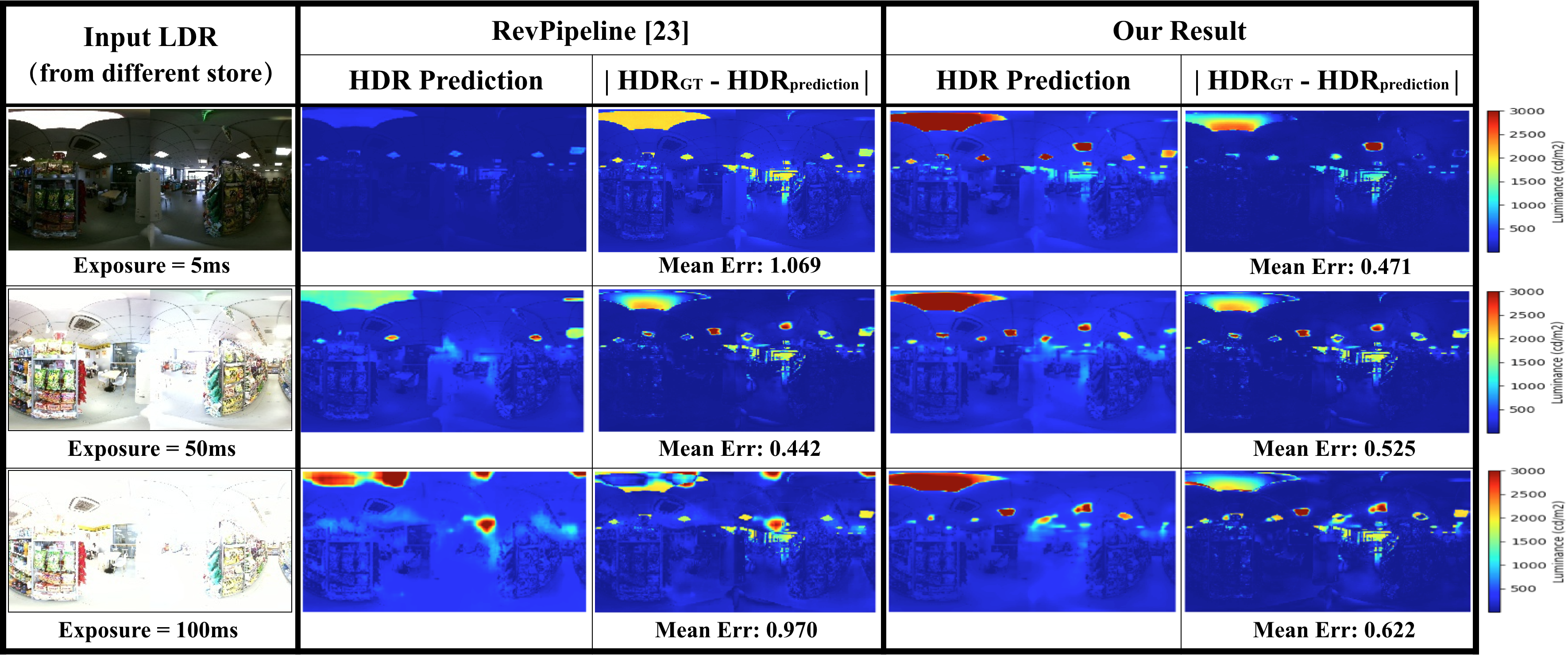}
    \caption{Visualization of cross-domain study.}
    \end{subfigure}
    \vspace{-3mm}
    \caption{HDR generation consistency evaluation under different exposure settings. Our HDRs are more consistent comparing against ground truth HDR than \cite{liu2020single} in both same and cross-domain tests. Note that we visualize the performance of HDR by its luminance map and corresponding false color map, independent of tone-mapping techniques \cite{ledda2005evaluation,liang2018hybrid} for HDR display.}
    \label{fig:comparison}
\end{figure*}

\subsection{Results of Cross-domain Study}

\begin{table}[!t]
\centering
\vspace{-4mm}
{\small
\begin{tabular}{|c | c| c| c| c|} 
 \hline
  Methods & \scriptsize{HDR-VDP}$\uparrow$ & \scriptsize{Mean Std}$\downarrow$ & \scriptsize{25\% Acc}$\uparrow$ & \scriptsize{10\% Acc}$\uparrow$ \\ 
 \hline\hline
 \multicolumn{5}{|c|}{\textbf{Multi-shot method} \cite{ward2016anyhere}}\\
 \hline
 \scriptsize{Before Calibration} & --- & --- & 27.08\% & 7.19\% \\
 \scriptsize{After Calibration} & --- & --- & 65.71\% & 34.52\%\\
 \hline
 \multicolumn{5}{|c|}{\textbf{Single-shot methods}}\\
 \hline
 HDRCNN\cite{eilertsen2017hdr} & 64.2955 & 0.2935 &  8.16\% & 2.89\% \\
 DrTMO\cite{endo2017deep} & 69.4140 & 0.2855 & 13.08\% & 2.02\% \\ 
 RevPipeline \cite{liu2020single} & 69.8549 & 0.6145 & 21.07\% & 7.87\% \\ 
 Ours &  70.0474 & 0.2618 & 71.82\% & 38.25\% \\ 
 Ours+ &  70.1064 & \textbf{0.2571} & \textbf{72.36\%} & \textbf{40.64\%} \\ 
 \hline
\end{tabular}
}
\vspace{-2mm}
\caption{Cross-domain quantitative comparison on HDR images with existing methods.}
\label{table:diff}
\end{table}

We also conduct experiments that training and testing share disjoint stores to verify the generalizability of the proposed method in the case of slightly domain shift. We randomly select 5 stores as training, 1 as validation and the rest 2 as testing. The results are shown in an average of five repeated trials with different random seeds.

As shown in Table~\ref{table:diff}, as expected, the learning-based state-of-the-art single-shots methods without illuminance supervision achieved similar results when applied cross-domain. Even after calibration with the optimal scaling factor, the illuminance value derived from calibrated HDR image still deviates from the ground truth measured by illuminometer. Due to domain shift, our method also suffers from an accuracy drop in illuminance estimation but still outperforms others in all metrics. To further boost the performance, we pretrain the backbone network without illuminance supervision with randomly cropped image patches, and then finetune with the proposed method. This is named \textsc{Ours+} in Table \ref{table:diff} and it can be viewed as a data augmentation trick to slightly improve the performance over the proposed method.

In Figure \ref{fig:comparison}, we compare our method with the state-of-the-art RevPipeline \cite{liu2020single} in visualizing the HDR prediction results and false color maps against ground truth HDR for LDRs with different exposures. More results will be shown in supplementary material.


\subsection{Ablation Study}

\begin{figure*}
    \vspace{-4mm}
    \centering
    \includegraphics[width=\linewidth,height=2.5cm]{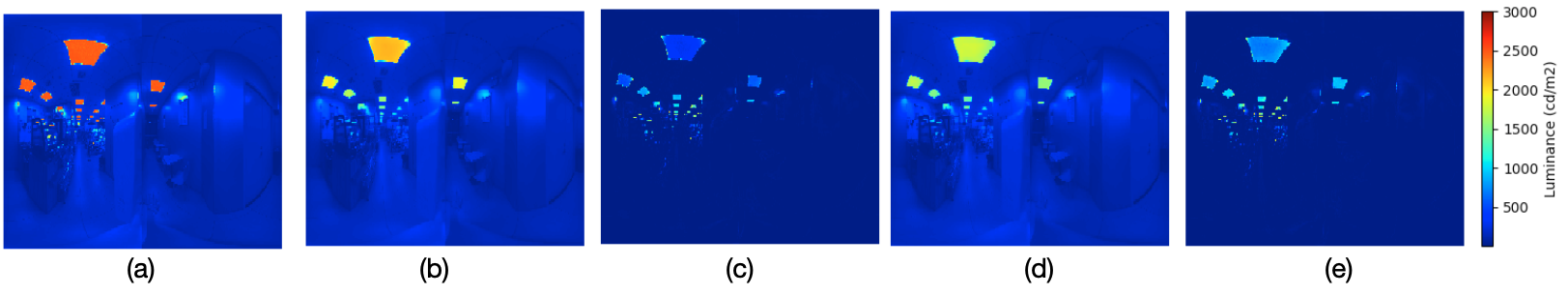}
    \vspace{-6mm}
    \caption{(a) Ground truth HDR luminance map; (b,c) Predicted HDR luminance map and false color map of \textsc{Ours$\backslash$Integral}. The mean of error map is 0.161; (d,e) Predicted HDR luminance map and false color map of \textsc{Ours}. The mean of error map is 0.823.}
    \label{fig:ablation}
    \vspace{-4mm}
\end{figure*}

\noindent
\textbf{Function of Illuminance Supervision.} One of the key designs in our method is incorporating illuminance estimation as a penalty term for HDR image reconstruction. Combined with the reconstruction loss against the ground-truth HDR image, the illuminance supervision makes the reconstructed HDR image more physically plausible. For both same-domain and different-domain study, as shown in Table~\ref{table:ablation}, we observed that by eliminating the illuminance supervision as the integral term in Equation~\ref{eq:loss_total}, the accuracy of illuminance estimation drops dramatically but the metric of HDR-VDP is slightly better, also evidenced in Figure \ref{fig:ablation}. These can be explained by the fact that ground-truth HDR images from the multi-shot method are still far from perfect. By adding the integral term, the learning process drives the HDR reconstruction toward a more physically plausible goal, not only just as close as the ground-truth.


\begin{table}[!t]
\centering
{\small
\begin{tabular}{|c|c|c|c|c|} 
 \hline
 \multicolumn{1}{|c|}{\textbf{Terms}} & \multicolumn{4}{|c|}{\textbf{Metrics}}\\
 \hline
 & \scriptsize{HDR-VDP}$\uparrow$ & \scriptsize{Mean Std}$\downarrow$ & \scriptsize{25\% Acc}$\uparrow$ & \scriptsize{10\% Acc}$\uparrow$ \\ 
 \hline\hline
 \multicolumn{5}{|c|}{\textbf{Same-domain study}}\\
 \hline
  Ours$\backslash$Integral & \textbf{72.4706} & \textbf{0.0856} &  72.72\% & 35.28\% \\ 
  Ours$\backslash$Exposure & 71.5549 & 0.1974 & 99.30\% & 86.56\% \\ 
  Ours & 72.1717 & 0.0959 & \textbf{99.98\%} & \textbf{97.95\%} \\
 \hline
 \multicolumn{5}{|c|}{\textbf{Different-domain study}}\\
 \hline
  Ours$\backslash$Integral & \textbf{70.1737} & 0.2649 & 60.02\% & 29.42\%  \\
  Ours$\backslash$Exposure & 69.8093 & 0.2778 & 69.68\% & 36.65\%  \\
  Ours & 70.0474 & \textbf{0.2618} & \textbf{71.82\%} & \textbf{38.25\%} \\
 \hline
\end{tabular}
}
\vspace{-2mm}
\caption{Ablation study. The function of illuminance supervision and exposure encoding.}
\label{table:ablation}
\vspace{-4mm}
\end{table}

\noindent
\textbf{Function of Exposure Encoding.} In our method, the parameter of exposure value which may be the most important camera configuration parameter for the investigated task is considered as an auxiliary input and specific network designs are made to fuse the exposure feature words with the image-level features. As shown in Table~\ref{table:ablation}, without explicit exposure encoding, the Mean-Std metric is both higher for the same and cross domain study, which indicates that explicit exposure encoding can make the reconstructed HDR images of LDR bracket more consistent, as a result, to improve the robustness of the learning framework. For the metric of HDR-VDP and illuminance estimation accuracy, the exposure-conditioned design also brings slight improvements by regularizing the HDR reconstruction. 

\noindent
\textbf{Function of HDR Generation for Illuminance Estimation.} 
We have demonstrated that our method can reconstruct HDR images which are physically plausible, \ie, with the potential to estimate the illuminance accurately. One would question whether accurate illuminance could be estimated from LDR image directly by learning a mapping from LDR image to the ground truth illuminance reading, without the intermediate process of HDR reconstruction. 

With the same amount of data in our previous experiments, we adopt the popular Spherical CNNs \cite{cohen2018spherical} designed for equirectangular images to learn a direct mapping from LDRs to the ground truth illuminance readings, without the intermediate HDR reconstruction. As shown in Table~\ref{table:ldr2illuminance}, the illumiance estimation accuracy for the direct mapping method are not good as our method, which indicates the importance of having HDR image reconstruction as an intermediate step. We argue that with HDR reconstruction in between, the illuminance estimation process is unrolled following a physic-based process. Hence the learning is easier with better generalization. 

\begin{table}[!t]
\centering
\begin{tabular}{|c|c|c|} 
 \hline
 \multicolumn{1}{|c|}{\textbf{Terms}} & \multicolumn{2}{|c|}{\textbf{Metrics}}\\
 \hline
 & 25\% Acc$\uparrow$ & 10\% Acc$\uparrow$ \\ 
 \hline\hline
 \multicolumn{3}{|c|}{\textbf{Same-domain study}}\\
 \hline
  LDR2Illuminance &  90.92\% & 52.01\% \\ 
  Ours & \textbf{99.98\%} & \textbf{97.95\%} \\
 \hline
 \multicolumn{3}{|c|}{\textbf{Cross-domain study}}\\
 \hline
  LDR2Illuminance  & 63.21\% & 32.99\%  \\
  Ours & \textbf{71.82\%} & \textbf{38.25\%}\\
 \hline
\end{tabular}
\vspace{-2mm}
\caption{Ablation study. The function of HDR reconstruction as an intermediate process to estimate the illuminance.}
\label{table:ldr2illuminance}
\vspace{-4mm}
\end{table}

\section{Conclusion} \label{sec:con}
In this work, we consider a new problem: reconstructing physically plausible HDR from a single-shot. We present a new learning-based method that generates HDR images not only being visually appealing but also physically correct. To achieve this goal, we introduce two new constraints: (1) illuminance correctness - that the output HDR should agree with the physical illuminance measure at the imaging location; (2) multi-exposure consistency - different LDR images with only exposure difference should output the same HDR. We collect a large dataset that couples the images with the illuminance readings to evaluate our method and could help facilitate research in this area.

In our experiments, if the network is trained and tested in the same store with the same imaging device, 99.98\% of the locations measured by our method are within an absolute error of 25\% from the ground truth measurement. This is an acceptable performance for the interior design industry. Although our cross-domain performance is not as good as the same-domain results, it achieves better accuracy than the classic multi-exposure HDR generation method without delicate cross-imaging-device calibration and response curve scale normalization. Last but not the least, although the exposure settings as inputs can help improve the HDR reconstruction, our network does not require to train and infer with it. When they are unknown, as shown in Table 4, our method can still achieve very competitive results.

We believe the proposed method leads a promising path to direct physical property measurement. Future directions include but not limited to extending the HDR scope from the studied spherical panoramas to general images captured by less than 180-degree FOV cameras. One possible solution is by careful masking on the illuminometer that corresponds to the camera FOV. Moreover, the cropped version of our reconstructed HDR results could be used as training data for general single-shot HDR reconstruction task to achieve physically plausible HDRs.

{\small
\bibliographystyle{ieee_fullname}
\bibliography{egbib}
}

\end{document}